\documentclass[conference]{Format/IEEEtran}
\IEEEoverridecommandlockouts

\usepackage{acro}
\usepackage[utf8]{inputenc}
\usepackage[ruled,vlined,linesnumbered]{algorithm2e}
\usepackage{amsfonts, amsmath, amssymb}
\usepackage{array}
\usepackage{bm}
\usepackage{cite}
\usepackage{color}
\usepackage{xcolor}
\usepackage{float}
\usepackage{graphicx}
\usepackage{subfig}
\usepackage{hyperref}
\usepackage{url}
\usepackage{multirow}
\usepackage{booktabs}
\usepackage{siunitx}

\usepackage{dblfloatfix} 
\usepackage[ancient]{flushend} 

\title{Nukhada USV: a Robot for Autonomous Surveying and Support to Underwater Operations}

\author{
    {\`E}ric Pairet$^1$, Simone Span{\`o}$^2$, Nikita Mankovskii$^1$, Paolo Pellegrino$^1$\\
    Igor Zhilin$^1$, Jeremy Nicola$^1$, Francesco La Gala$^2$, Giulia De Masi$^1$
    \\[0.15cm]
    $^1$The Technology Innovation Institute (TII), Abu Dhabi, United Arab Emirates \\
    \{\textit{eric.pairet, nikita.mankovskii, paolo.pellegrino, igor.zhilin, jeremy.nicola, giulia.demasi}\}\!\atsign tii.ae \\[0.15cm]
    $^2$SpinItalia, Roma, Italy \\
    \{\textit{s.spano, f.lagala}\}\!\atsign spinitalia.com
    \thanks{All authors contributed equally to this work.}
}

\newcommand\atsign{@}

\newcommand*{\sref}[1]{Section~\ref{#1}}

\newcommand*{\fref}[1]{Figure~\ref{#1}}

\definecolor{purple}{RGB}{106,13,173}

\DeclareAcronym{1D}{
  short = 1D,
  long  = one-dimensional
}
\DeclareAcronym{2D}{
  short = 2D,
  long  = two-dimensional
}
\DeclareAcronym{3D}{
  short = 3D,
  long  = three-dimensional
}
\DeclareAcronym{ADCP}{
  short = ADCP,
  long  = acoustic Doppler current profiler
}
\DeclareAcronym{AI}{
  short = AI,
  long  = artificial intelligence
}
\DeclareAcronym{AIS}{
  short = AIS,
  long  = automatic identification system
}
\DeclareAcronym{AUV}{
  short = AUV,
  long  = autonomous underwater vehicle
}
\DeclareAcronym{BMS}{
  short = BMS,
  long  = battery management system
}
\DeclareAcronym{CAN}{
  short = CAN,
  long  = controller area network
}
\DeclareAcronym{DoF}{
  short = DoF,
  long  = degree of freedom,
  long-plural-form = degrees of freedom
}
\DeclareAcronym{DVL}{
  short = DVL,
  long  = Doppler velocity logger
}
\DeclareAcronym{FLS}{
  short = FLS,
  long  = forward looking system
}
\DeclareAcronym{GNSS}{
  short = GNSS,
  long  = global navigation satellite system
}
\DeclareAcronym{HRI}{
  short = HRI,
  long  = human-robot interaction
}
\DeclareAcronym{IMU}{
  short = IMU,
  long  = inertial measurement unit
}
\DeclareAcronym{INS}{
  short = INS,
  long  = inertial navigation system
}
\DeclareAcronym{LED}{
  short = LED,
  long  = light-emitting diode
}
\DeclareAcronym{LiDAR}{
  short = LiDAR,
  long  = light detection and ranging
}
\DeclareAcronym{LMS}{
  short = LMS,
  long  = least mean squares
}
\DeclareAcronym{NMSE}{
  short = NMSE,
  long  = normalised mean squared error
}
\DeclareAcronym{NN}{
  short = NN,
  long  = neural network
}
\DeclareAcronym{MBZIRC}{
  short = MBZIRC,
  long  = Mohamed Bin Zayed international robotics challenge
}
\DeclareAcronym{MEMS}{
  short = MEMS,
  long  = Micro-electro-mechanical system
}
\DeclareAcronym{OMPL}{
  short = OMPL,
  long  = Open Motion Planning Library
}
\DeclareAcronym{PD}{
  short = PD,
  long  = proportional-derivative
}
\DeclareAcronym{ROS}{
  short = ROS,
  long  = robot operating system
}
\DeclareAcronym{RTK}{
  short = RTK,
  long  = real-time kinematic
}
\DeclareAcronym{UAV}{
  short = UAV,
  long  = unmanned aerial vehicle
}
\DeclareAcronym{USBL}{
  short = USBL,
  long  = ultra-short baseline
}
\DeclareAcronym{USV}{
  short = USV,
  long  = unmanned surface vehicle
}
\DeclareAcronym{UUV}{
  short = UUV,
  long  = unmanned underwater vehicles
}
\begin{document}
    \maketitle

    \begin{abstract}
    The Technology Innovation Institute in Abu Dhabi, United Arab Emirates, has recently finished the production and testing of a new \acl{USV}, called Nukhada, specifically designed for autonomous survey, inspection, and support to underwater operations. This manuscript describes the main characteristics of the Nukhada USV, as well as some of the trials conducted during the development.
\end{abstract}
    \section{INTRODUCTION \label{sec:introduction}}
As the era of space exploration is taking off, we still know very little about our oceans. It is estimated that more than 80\% of the marine environment remains unexplored. Yet, the few we know about it has made evident the relevance of oceans in oxygen production, climate regulation, and serving a key role in our energy and economic infrastructures. Unravelling the mysteries of such an environment, as well as preserving it for a sustainable future, 
is essential for a better world.

\begin{figure}[t!]
    \centering
    \includegraphics[width=0.9\columnwidth]{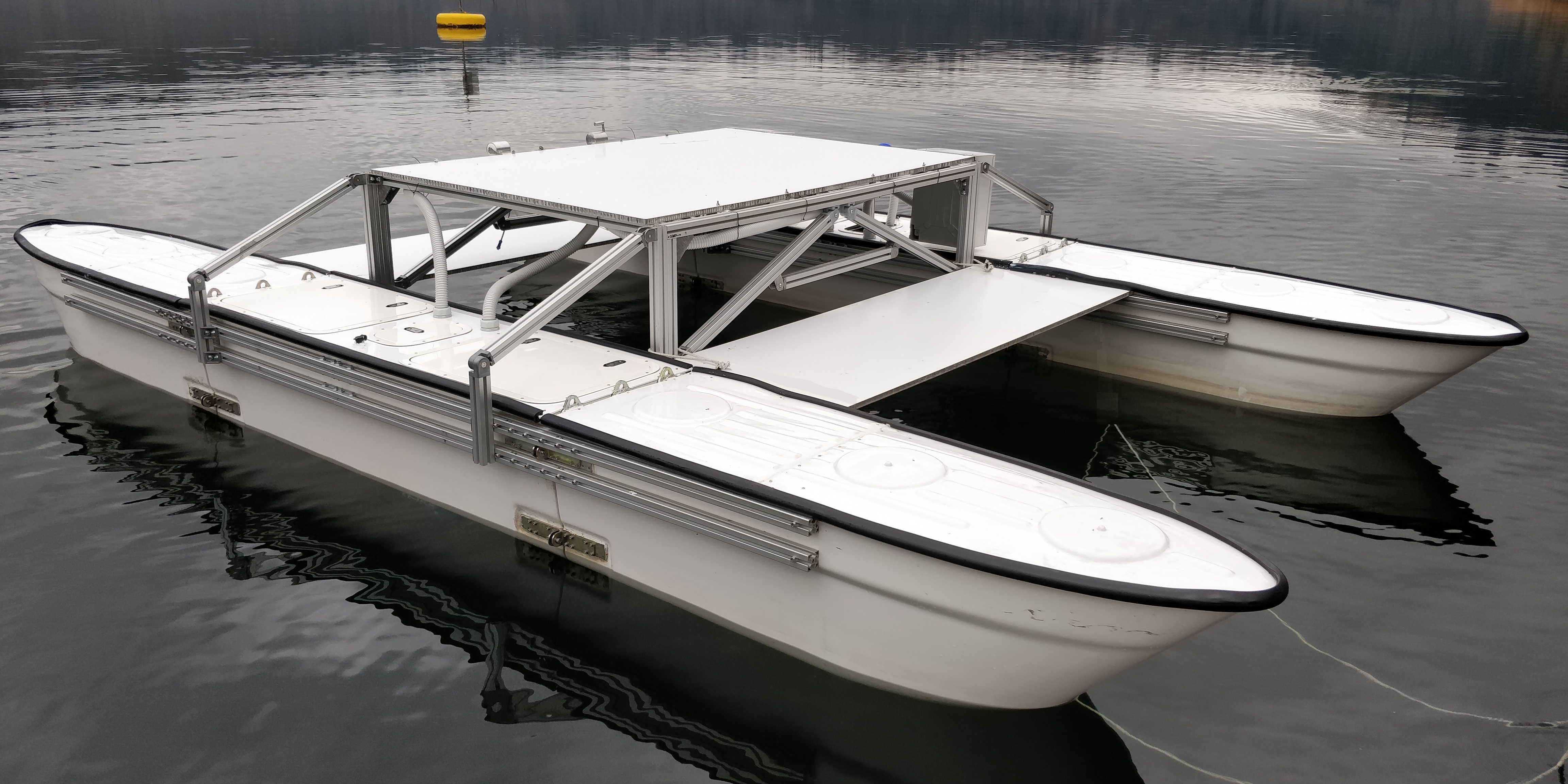}
    \caption{Nukhada~USV, TII's new USV for autonomous survey, inspection, and support to underwater robotic operations.}
    \label{fig:nukhada}
    \vspace{-0.4cm}
\end{figure}

The Technology Innovation Institute (TII) in Abu Dhabi, United Arab Emirates, is committed to developing the most advanced, disruptive technological innovations to solve society’s greatest challenges. In the context of understanding and preserving our oceans, TII is developing marine robotic platforms that can support a variety of missions, such as cooperative exploration, human-made structure inspection and maintenance, and among many others, search and rescue. This manuscript presents the first robot of TII's fleet of marine vehicles, the Nukhada~USV (see \fref{fig:nukhada}). We believe that this paper, as previous publications of similar nature, e.g.,~\cite{ribas2011girona,carreras2015testing,willners2021from}, as well as a recent review on design choices of unmanned underwater vehicles~\cite{neira2021review}, is of relevance to the robotics community at large as it can serve as inspiration for the hardware and software design of future marine platforms.

\begin{figure}[b!]
    \vspace{-0.45cm}
    \centering
    \includegraphics[width=0.6\columnwidth]{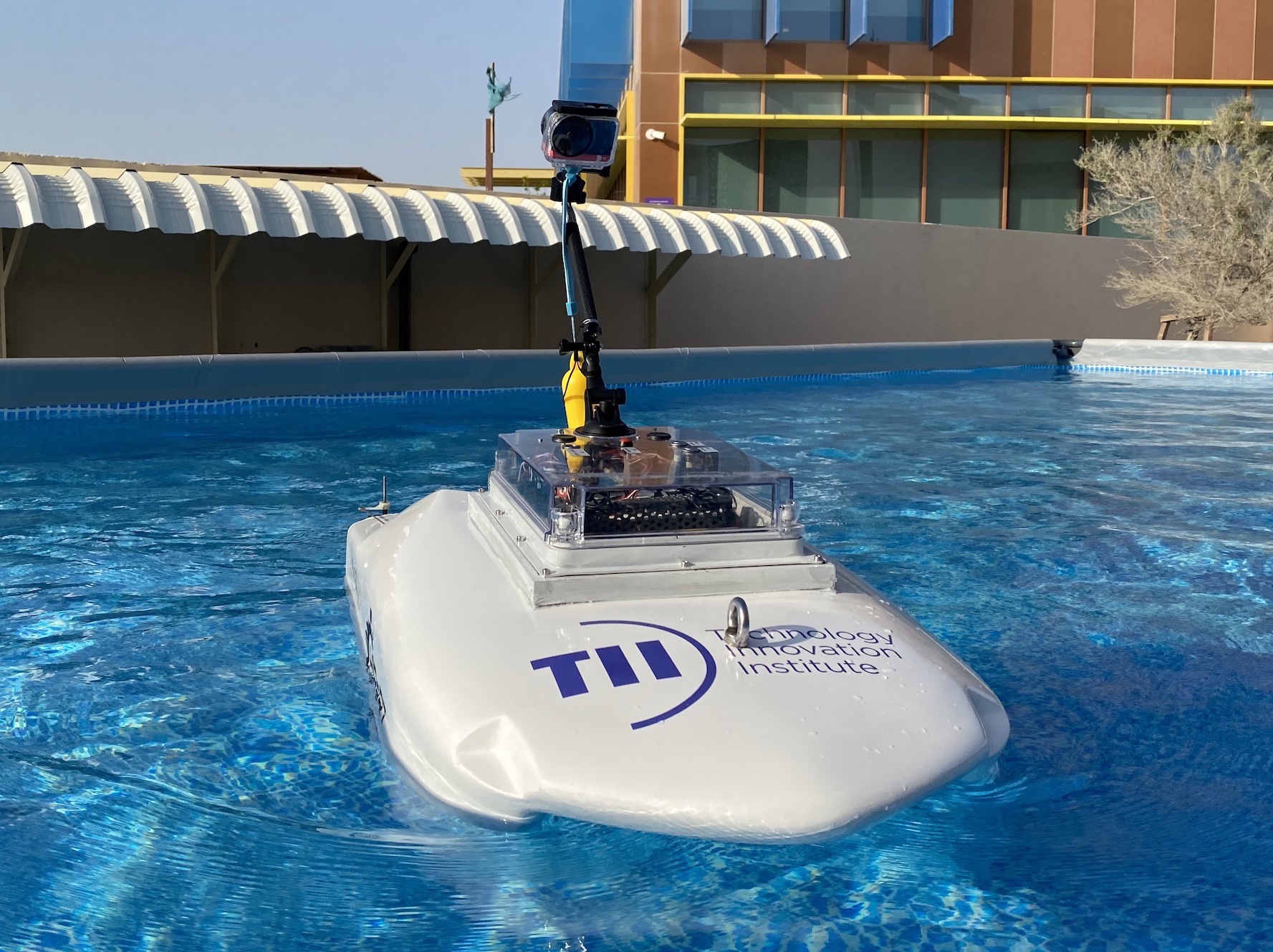}
    \caption{Nukhada-mini~USV, a 1:5 scaled version for steadily and safely progress from algorithmic prototyping to application deployment in real-world challenging scenarios.}
    \label{fig:nukhada_mini}
\end{figure}

A peculiarity of our Nukhada~USV is the parallel development of its 1:5 scaled version, the Nukhuda-mini~USV (see \fref{fig:nukhada_mini}).
The scaled vehicle is equipped with the same sensor technology (but generally with cheaper hardware alternatives) and the same software architecture.
That allows to include an intermediate step between simulations and real-world deployment, consisting of small-scale testing in controlled spaces, 
without the need to heavily modify algorithms when migrating between vehicles.
Given the size of our full-scale Nukhada~USV, having an easier to deploy, scaled version is particularly relevant as most controlled spaces are water or waves tanks of reduced space, and testing in open waters with the final platform requires more attention to safety.

The remainder of this article focuses on the full-scale Nukhada USV as follows.
\sref{sec:mechanical_design} presents details on the vehicle's mechanical design. \sref{sec:power_system} describes the system that powers up the vehicle, while \sref{sec:computer_architecture} focuses on the hardware and software that enable the autonomy architecture. \sref{sec:payload_equipment} describes the basic sensor suite, while \sref{sec:results} exemplifies payloads for some mission-specific under development. Finally, \sref{sec:conclusion} concludes this manuscript.
    \begin{figure*}[b!]
    \centering
    \subfloat[Type-A: flat-ending]{\includegraphics[width=0.48\linewidth]{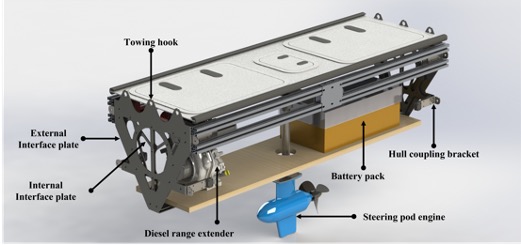} \label{fig:module_type_a}}%
    \quad
    \subfloat[Type-B: hydrodynamic-ending]{\includegraphics[width=0.478\linewidth]{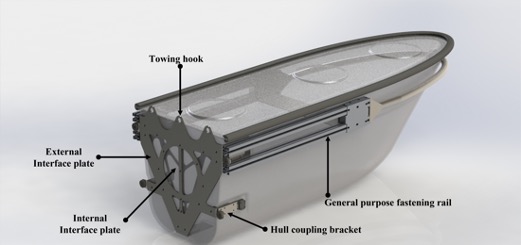} \label{fig:module_type_b}}%
    \\
    \begin{minipage}{0.69\textwidth}
        \centering
        \subfloat[Overview Nukhada~USV mechanical design]{\includegraphics[width=0.98\linewidth]{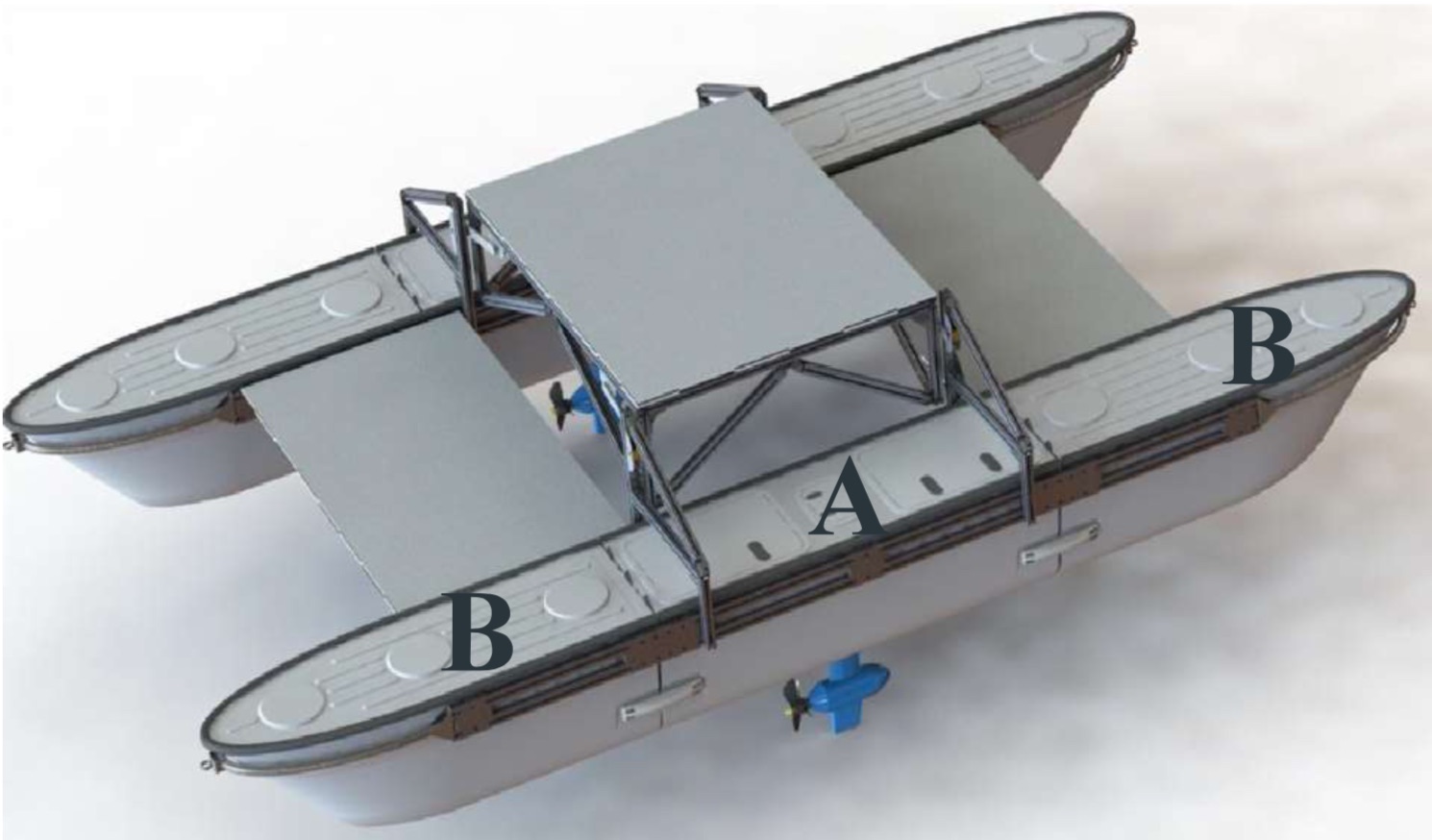} \label{fig:nukhada_sketch}}
    \end{minipage}
    \;%
    \begin{minipage}{0.28\textwidth}
        \centering
        \subfloat[Lower USV constitution]{\includegraphics[width=\linewidth]{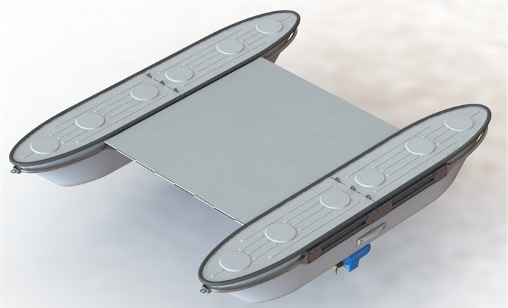} \label{fig:nukhada_variant_1}}
        \\
        \subfloat[Shorter USV constitution]{\includegraphics[width=\linewidth]{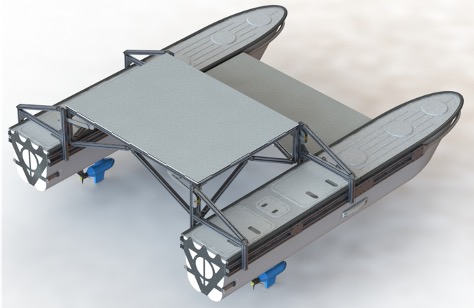} \label{fig:nukhada_variant_2}}
    \end{minipage}
    \caption{The Nukhada~USV is mechanically designed to promote modularity. The watertight hull modules (a)~type-A and (b)~type-B allow to arrange varied USV constitutions, as for example, (c)~the ${6 \times 3.5}$~metres Nukhada~USV, and ${4 \times 3.5}$~metres (d)~low-profile and (e)~shorter options. Larger USVs can be arranged by multiple intermediary Type-A modules.}
\end{figure*}

\section{MECHANICAL DESIGN \label{sec:mechanical_design}}

The Nukhada~USV is a ${6 \times 3.5}$~metres modular surface vessel (see~\fref{fig:nukhada_sketch}). It consists of symmetric double-ender hulls connected with large flat panels for the versatile installation of mission-specific payload. The \ac{USV}'s height is $1.4$~metres ($1$~metre over waterline), and its dry weight is $500$~kg. The hulls are made of several ${2}$~metres long fibreglass watertight modules coupled together; specifically, a type-A module (flat-ending module, see~\fref{fig:module_type_a}) in-between two type-B modules (hydrodynamic-ending modules, see~\fref{fig:module_type_b}).

The design of the vehicle paid special attention to modularity, allowing to easily swap parts for ease of maintenance with low downtime, while enabling for alternative constitutional USV arrangements (see examples \fref{fig:nukhada_variant_1} and \fref{fig:nukhada_variant_2}). The large flat panels between the two demi-hulls are made of honeycomb composite aluminium, providing a resistant but light area customisable to mission-specific needs. Additionally, general-purpose fastening rails all along the hulls allow for versatile and quick assembly of additional components. 

\begin{figure}[b!]
    \vspace{-0.9cm}
    \centering
    \subfloat[Operation mode: unfolded]{\includegraphics[height=2.4cm]{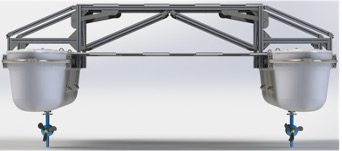}\label{fig:nukhada_mechanical_design_transport_unfolded}}
    \,
    \subfloat[Transport mode: folded]{\includegraphics[height=2.4cm]{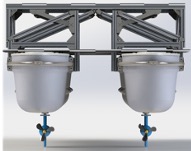}\label{fig:nukhada_mechanical_design_transport_folded}}
    \caption{The whole USV can be folded for ease of transportation into a compact ${6 \times 2}$~metres vehicle.}
    \label{fig:nukhada_mechanical_design_transport}
    \vspace{-0.15cm}
\end{figure}

The modules ease the transport, assembly and deployment of the USV. None of the modules nor platforms is longer than 2 metres, allowing for the whole vessel to be conveniently stored and shipped in pieces. The assembly of the vessel just requires basic tools, such as screwdrivers, wrenches and pliers, without the need for adhesives thanks to the fast and secure coupling techniques between hulls. Alternatively, the central part of the USV can be folded to narrow the vehicle's width (see \fref{fig:nukhada_mechanical_design_transport}), thus allowing for its transport on road by a car towed boat trailer. The presence of multiple hooking points in the stern, bow and the interface plates of each section of the catamaran allows for the towing and lifting of the vehicle.
    \section{POWER and PROPULSION SYSTEM \label{sec:power_system}}

\begin{figure}[b!]
    \centering
    \includegraphics[width=\columnwidth]{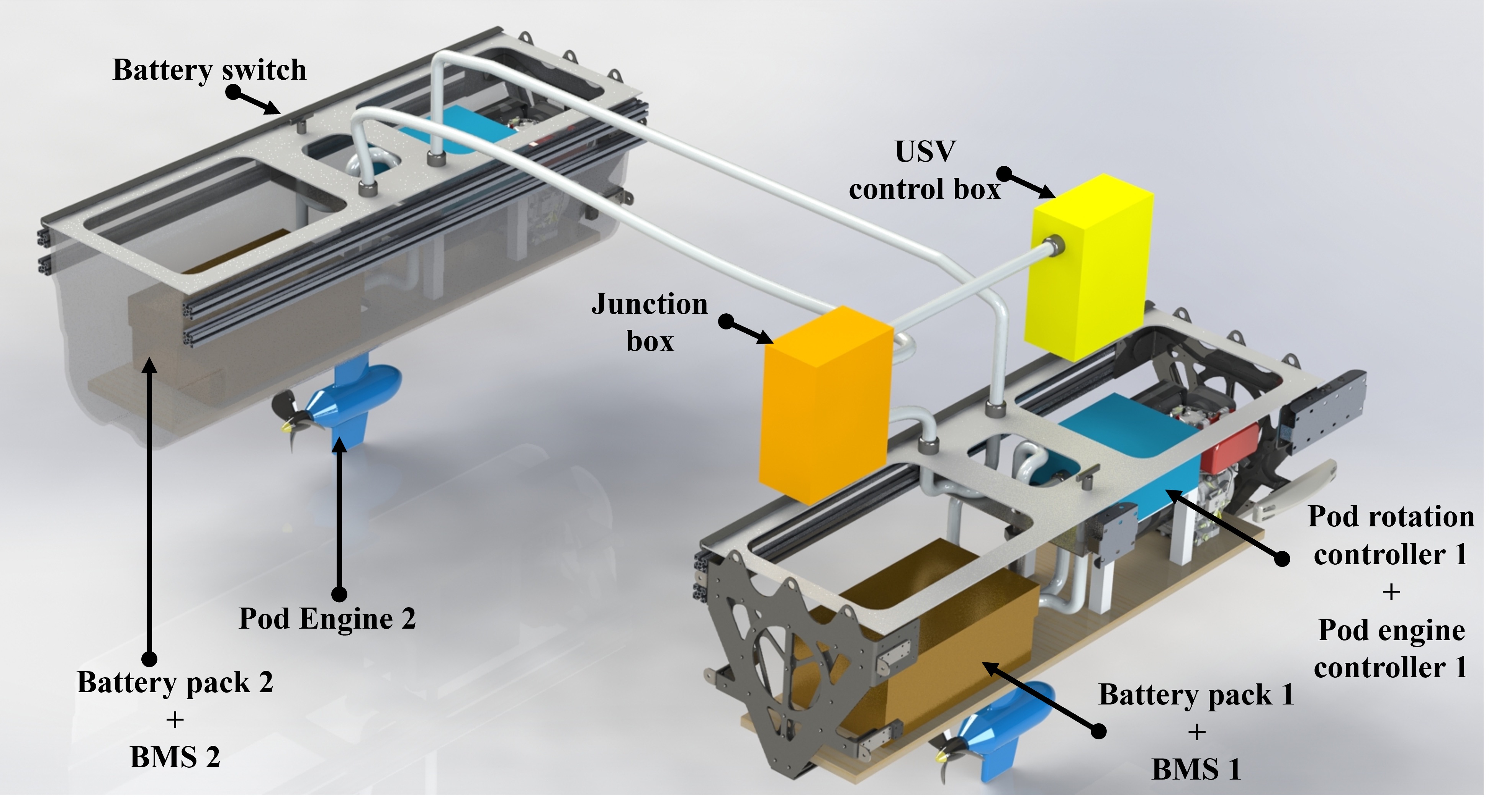}
    \caption{Power and propulsion system of the Nukhada~USV.}
    \label{fig:usv_essential_equipment}
\end{figure}

The catamaran is endowed with large operational range via a hybrid power system composed of batteries and diesel-electric generators (see \fref{fig:usv_essential_equipment}). Specifically, the central part of each demi-hull contains a battery pack of 16 LiFePO cells (3.2~V / 110~Ah) connected in series (providing a total of 5.6~kWh at 51.2~VDC nominal voltage), and a diesel range extender generator of 2.5~kW to recharge the battery pack. The battery packs in each demi-hull are connected in parallel in order to keep a balanced charge. Additionally, the amount of battery packs can be up to quadriplicated and the capacity of the fuel tank increased to extend the autonomy of the catamaran.

The charge of two standard battery packs takes $4$~hours, but only $1$~hour when employing a fast charger. Alternatively, $6$~square metres of solar panels providing a maximum power of $1.1$~kW can be installed to charge the batteries.

The propulsion system of the Nukhada USV counts on two electric engines located in the central part of each demi-hull. Electric engines allow for silent and environmentally friendly operations. The watertight pod engines integrate into the respective demi-hull with through-hull tubes. Each pod engine provides a maximum of $4.3$~kW on the shaft at $1350$~rpm. With the standard equipment and the base battery packs, the catamaran has an approximate $20$~nm autonomy with a cruise and maximum speed of $4$~kn and $8$~kn, respectively.

The pod engines are $\pm180$~degrees independently steerable with the catamaran both stationary and in motion, thus allowing for the catamaran's control in 3~\acp{DoF}: surge, sway and yaw. This propulsion setup enables high manoeuvrability in confined spaces but, optionally, four rudders can be added for unmatched manoeuvrability, as well as redundancy in case of failure; the catamaran can be manoeuvred only with one motor and one rudder. 

The electrical connection between the pod engine and its controller, the pod rotation actuator and its controller, and the batteries is conducted in a proper box inside each of the central hulls.
An on/off switch allows to manually control their 48~VDC connection. Additionally, two DC/DC converters with 48~VDC input, and 24~VDC and 12~VDC outputs allow powering all required hardware of the computer architecture and sensors.

    \section{COMPUTER ARCHITECTURE \label{sec:computer_architecture}}

\subsection{Hardware}
    
    All computational hardware for the autonomy architecture of the \ac{USV} is distributed into the \textit{control box} and the \textit{junction box}. These two compartments are installed onto the central module of the catamaran's right demi-hull holds (see \fref{fig:usv_essential_equipment}).
    
    The control box contains a modem/router Wi-Fi/4G to route data packets between the various Ethernet-connectable devices and allow remote communications. Also, this compartment has a dedicated computer for processing the point clouds of the onboard \ac{LiDAR} as obstacle information, and the main computer of the vehicle to perform point-to-point planning, local obstacle avoidance with the neighbouring obstacle information, and trajectory tracking. Both computers are inter-connected over Ethernet via the modem. A USB/CAN bus converter connected to the main computer enables communication over CAN bus.
    
    The junction box, besides the DC/DC converters of the power system (see \sref{sec:power_system}), contains a CAN bus-connected electronic board that acts as low-level controller. Specifically, this board translates velocity and heading setpoints, to commands for the pod engine and steer. Two additional displays indicate the status of the controllers of the pods. 
    
    Additional components to the catamaran's architecture can be connected either via Ethernet, Wi-Fi, or CAN bus. For instance, each battery pack \ac{BMS} is connected over CAN bus for essential health monitoring.
    
    \begin{figure}[b!]
        \centering
        \includegraphics[width=\columnwidth]{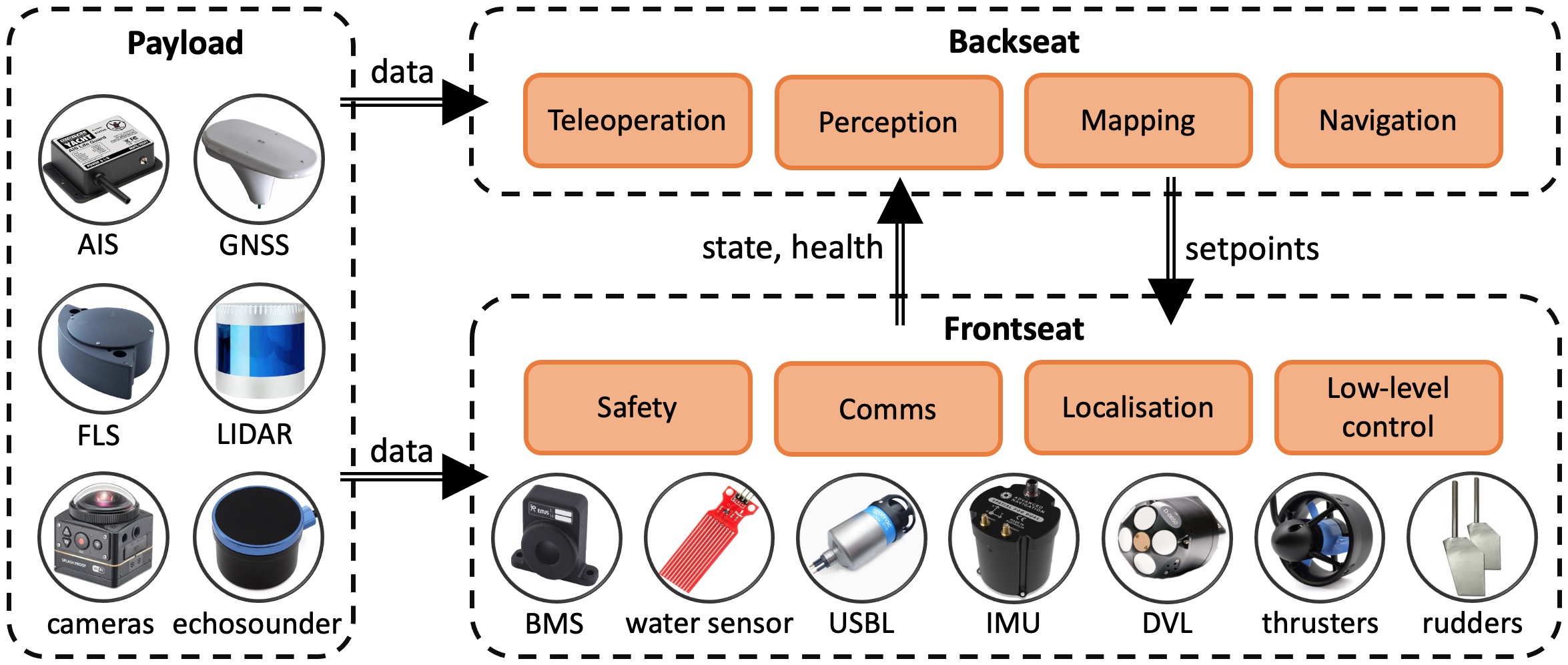}
        \caption{Nukhada's autonomy architecture.}
        \label{fig:nukhada_architecture}
    \end{figure}
    
\subsection{Software}
    The Nukhada~USV autonomy architecture consists of three main functional layers (see \fref{fig:nukhada_architecture}). Firstly, the \textit{frontseat} interfaces the essential actuators (i.e., thrusters and rudders), as well as interoceptive and exteroceptive sensors (see \sref{sec:payload_equipment}) to endow the robot with fundamental safety, communication, localisation and control routines. Secondly, the \textit{backseat} extends the vehicle's capabilities with more sophisticated algorithms for the robot autonomy, such as mapping, path planning and tracking, and among others, docking. Finally, the \textit{payload} includes some additional exteroceptive sensors (see \sref{sec:payload_equipment}) to feed the whole robot autonomy architecture with richer information about the surrounding environment. All these functional blocks run on Linux Ubuntu powered by the \ac{ROS}~\cite{quigley2009ros}.

    \section{SENSORS and PAYLOAD EQUIPMENT \label{sec:payload_equipment}}

The Nukhada~\ac{USV} is equipped with a selection of sensors to make it suitable to conduct autonomous surveying and support underwater operations. Namely, the onboard sensors can be divided into three sets: (a)~for obstacle detection above and under the water, (b)~for precise navigation even in the case of transient \acp{GNSS} loss, and (c)~for communication, localisation and collaboration with \acp{UUV}.

For detecting obstacles above the water, a total of five sensors are mounted on a mast to ensure maximum surrounding environment coverage: (a)~a \ac{LiDAR} for close to near high-resolution obstacle detection to navigate in challenging areas, (b)~a 360 degrees marine-grade radar for detecting ships, hard obstacles or coastal lines even at a great distance, (c)~a camera array arranged in a 360 degrees ring offering full surrounding visibility for detecting objects through machine learning methods, or offering a remote operator the possibility to visually assess the condition of the vehicle and potential damages after long operations, (d)~a forward-looking thermal camera, for reinforcing the visual obstacle detection in front of the vehicle, and (e)~a dual-band AIS receiver, for labelling detected obstacles and fusing additional information such as the heading and velocity.

To detect obstacles under the water, the vehicle is equipped with a volumetric 3D sonar mounted on a retractable pole in a forward-looking configuration. It allows for detecting any obstacle in the $120 \times 120$ degrees angle water column in front of the \ac{USV}, while measuring the bathymetry. The retractable pole allows removing the sensor from the water if need be.

For ensuring the precise localisation of the vehicle at all times, a miniature, dual-antenna \ac{RTK} \ac{GNSS} receiver aided \ac{MEMS} \ac{INS} provides accurate position, velocity, acceleration and orientation without relying on magnetometers. Moreover, a \ac{DVL} mounted on a retractable pole allows for lowering the inertial drift and maintaining a precise position estimation in case of \ac{GNSS} outages, as well as measuring current profiles when used in \ac{ADCP} mode.

For localising and communicating with \acp{UUV}, a GAPS M5 \ac{USBL} is mounted on a retractable pole.

Besides the permanent sensors that enable the vehicle for autonomous surveying and support underwater operations, the platform connecting both hulls serves as a multi-purpose dry payload area with a capacity of 600~kg for a range of additional sensors as well as for landing area of UAV, mounting of a LARS for deploying other vehicles or installing additional solar panels. A wet payload area below the waterline allows for the integration of underwater sensing.
    \section{DEVELOPMENT and RESULTS \label{sec:results}}

\begin{figure}[h!]
    \centering
    \includegraphics[width=0.8\columnwidth]{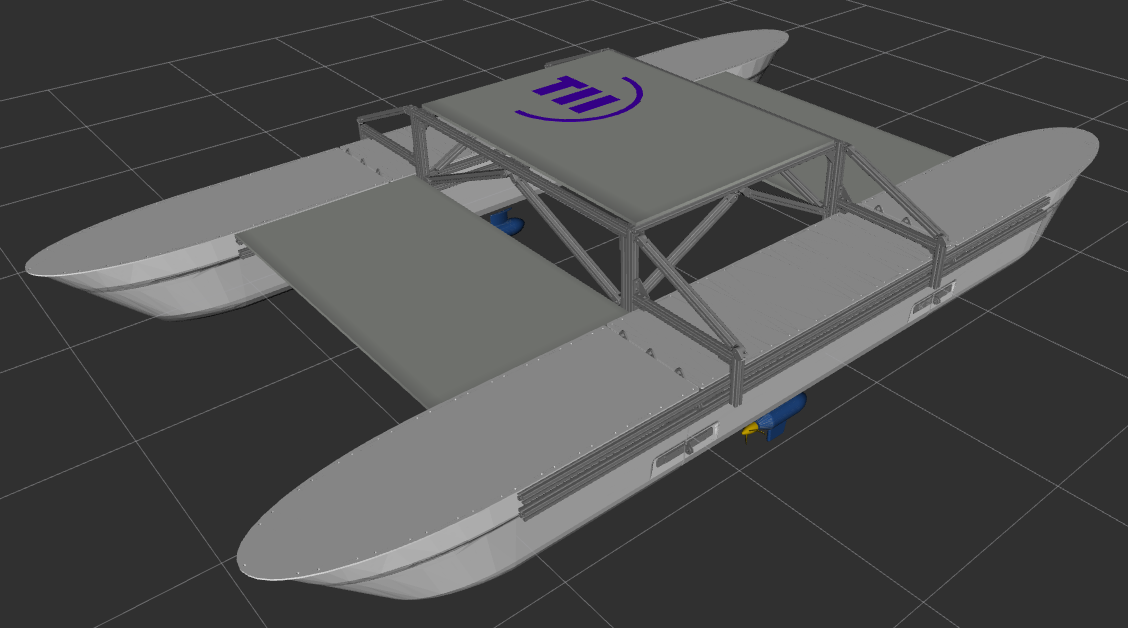}
    \caption{URDF model of the Nukhada \ac{USV} for simulations.}
    \label{fig:nukhada_urdf}
\end{figure}

The development and testing of our Nukhada \ac{USV} have been supported by its scaled version Nukhada-mini \ac{USV}, and them both with appropriate fully \ac{ROS}-integrated simulations (see \fref{fig:nukhada_urdf} for an example of the simulated Nukhada \ac{USV}). That journey supports our vision on steadily and safely progressing from hardware and algorithmic prototyping, to application deployment in real-world challenging scenarios. While initial testing on the Nukhada-mini USV finished in July 2021, the Nukhada USV is currently at the final stage of the in-water testing phase in which the vehicle is incrementally tested for longer mission times.

\subsection{Bathymetry with Nukhada-mini USV}

\begin{figure*}[t!]
        \centering
        \subfloat[Nukhada-mini \ac{USV} during the coverage survey.]{\includegraphics[width=0.342\linewidth]{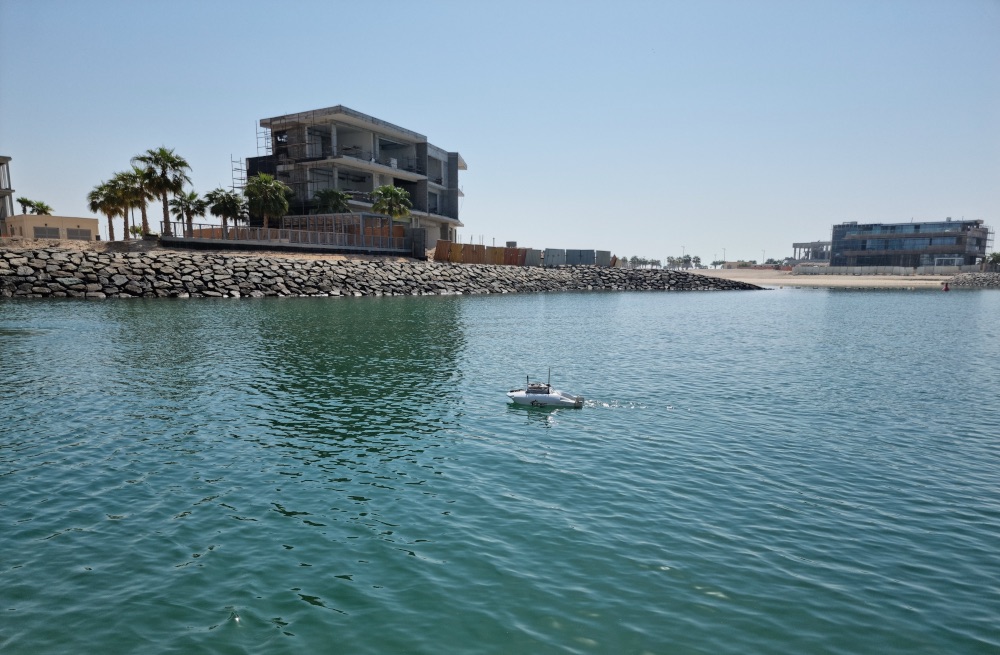} \label{fig:nukhada-mini_bathymetry_vehicle}}%
        \subfloat[Mission's planned coverage path.]{\includegraphics[width=0.324\linewidth]{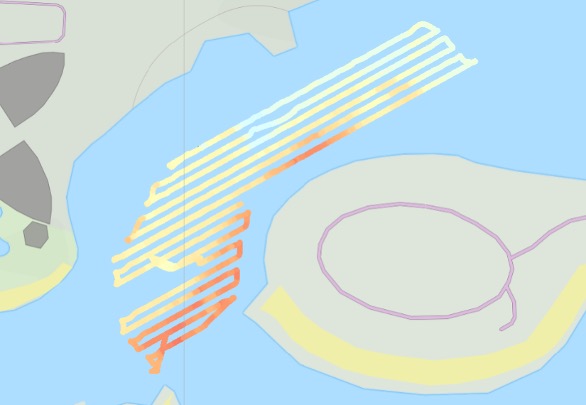} \label{fig:nukhada-mini_bathymetry_map_earth}}%
        \subfloat[Al Bateen bathymetric map over satellite image.]{\includegraphics[width=0.324\linewidth]{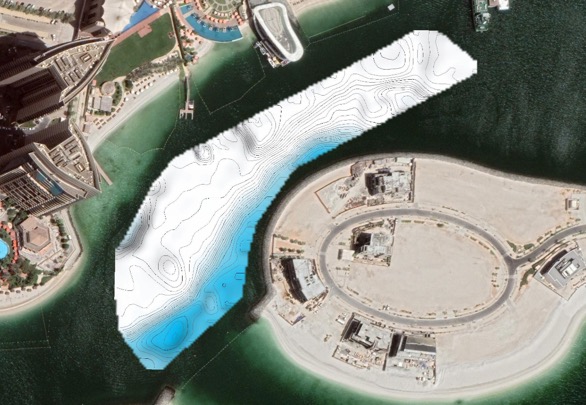} \label{fig:nukhada-mini_bathymetry_reefmaster}}%
        \caption{Bathymetry with Nukhada-mini USV in Al Bateen, Abu Dhabi, United Arab Emirates.}
    \end{figure*}

    
    To test the Nukhada-mini \ac{USV} as a whole, from its hardware to the software stack for autonomy, we executed a bathymetric survey of a large area in the creek of Al Bateen in Abu Dhabi, United Arab Emirates (see \fref{fig:nukhada-mini_bathymetry_vehicle}). The vehicle was equipped with an \ac{RTK} \ac{GNSS} and a single-beam echosounder from BlueRobotics in a down-looking configuration to function as altimeter. The borders of the area to inspect within the creek were specified by GPS coordinates, defining an area to be inspected of almost $31{,}000$ square meters. We used the coverage planning algorithm in~\cite{coverageplanner} to compute a lawnmower mission with $10$ meters distance between sweeps. The computed path, which can be seen in \fref{fig:nukhada-mini_bathymetry_map_earth}, for the coverage mission had an approximate length of $3.3$ kilometres, which the vehicle executed in about $1.5$ hours. The altimetry data collected with the single-beam echosounder during the survey was processed with the ReefMaster software\footnote{\url{https://reefmaster.com.au/}} to create the bathymetric map displayed in \fref{fig:nukhada-mini_bathymetry_reefmaster}. The depth of the surveyed area ranged between $3$ and $6.5$ metres.
    
\subsection{Manoeuvres with Nukhada USV}
    
    
    \begin{figure}[b!]
        \vspace{-0.3cm}
        \centering
        \subfloat[Outboard engine configuration.\break {\scriptsize Credit:~\url{https://youtu.be/K4YNAUET3Pc}}]{\includegraphics[width=0.49\columnwidth]{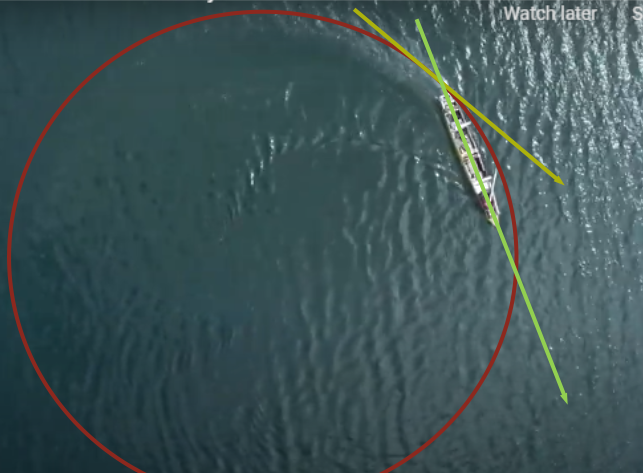}\label{fig:spiral_test_comparison_common}}
        \,
        \subfloat[Nukhada USV.]{\includegraphics[width=0.49\columnwidth]{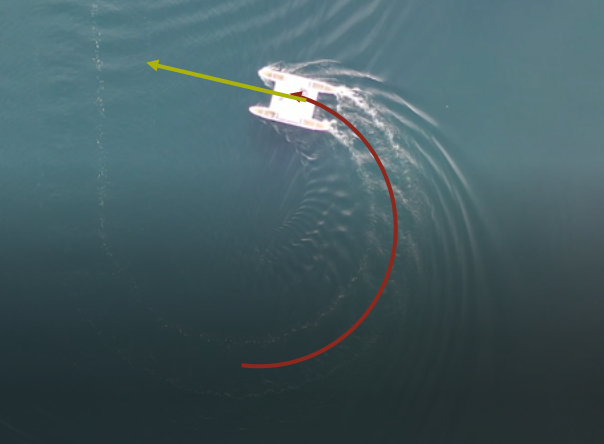}\label{fig:spiral_test_comparison_nukhada}}
        \caption{Behaviour comparison between a vessel with outboard engine setup and the Nukhada USV with a symmetrical setup.}
        \label{fig:spiral_test_comparison}
    \end{figure}
    
    To validate the behaviour of the Nukhada \ac{USV} and to estimate its hydrodynamic coefficients, we conducted several manoeuvres in Lake Nemi, Lazio, Italy. Among them, we detail two: fixed-radius turning and the spiral manoeuvre.
    
    The fixed-radius turning manoeuvre highlights the advantages of the Nukhada~USV's design. With the purpose of obtaining high manoeuvrability and stability, we adopted a symmetrical constitution not only on its geometrical structure (see \sref{sec:mechanical_design}), but also in the mass distribution and the propulsion system (see \sref{sec:power_system}), as opposed to classic outboard engine setups. \fref{fig:spiral_test_comparison} illustrates the contrasting performance of a vessel with classic outboard engine and the Nukhada~USV while executing the same fixed-radius turning manoeuvre (red trajectory). As it can be seen in \fref{fig:spiral_test_comparison_common}, the classic engine configuration leads to a considerable difference between the course (yellow line) and heading (green line) angles due to the action of the hydrodynamic forces raising from the non-symmetrical design. Instead, as shown in \fref{fig:spiral_test_comparison_nukhada}, the difference between the course and the heading angles is negligible for the symmetrical design of the Nukhada~USV. In all, the whole symmetry of the catamaran leads to symmetrical hydrodynamic forces, thus simplifying the vehicle's dynamical model and its controllability.


    
    
    
\begin{figure}[b!]
        \centering
        \subfloat[Propellers' command during the spiral tests.]{\includegraphics[width=1.0\linewidth]{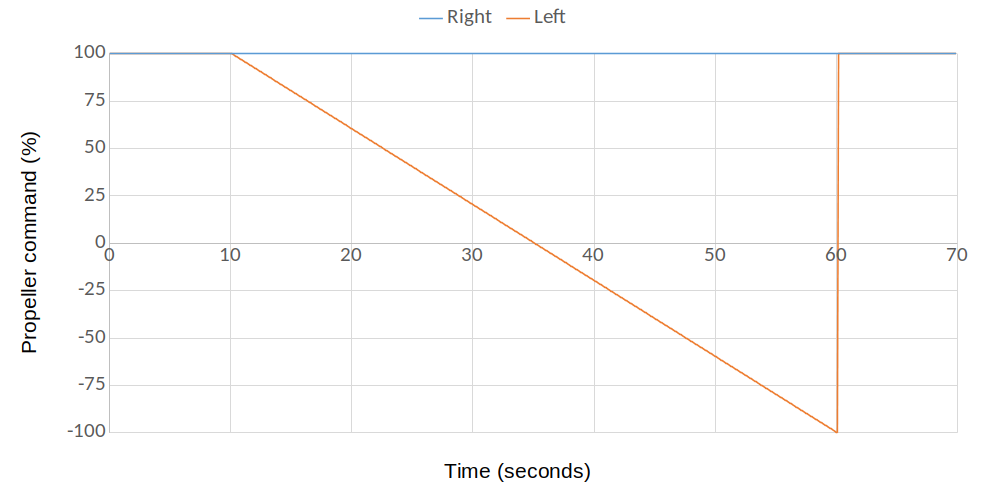} \label{fig:nukhada_tests_spiral_propellers_cmd}}%
        \\
        \subfloat[Top-view of the spiral executed by the vessel. Note the waves hinting the vessel's trajectory.]{\includegraphics[width=0.9\linewidth]{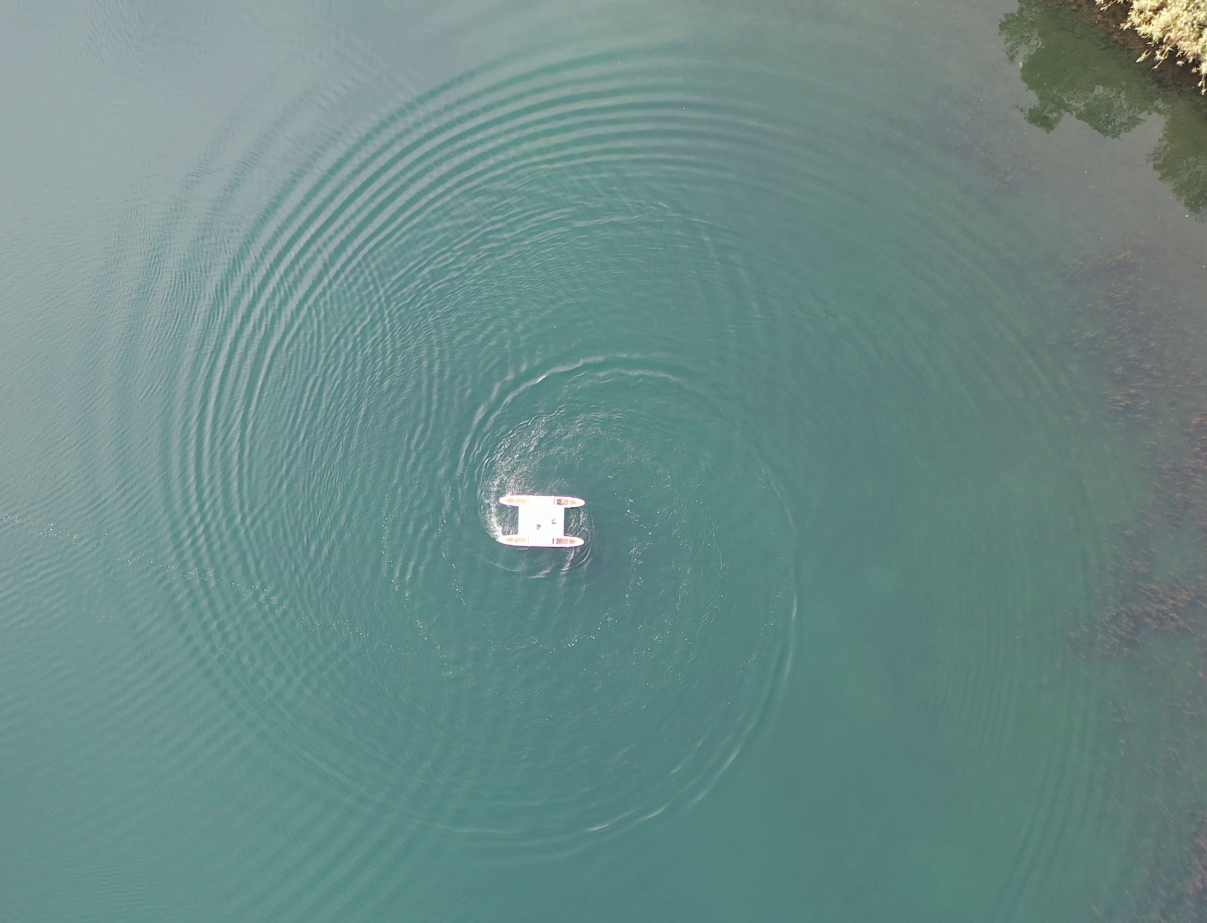} \label{fig:nukhada_tests_spiral}}%
        \caption{Spiral test executed with the Nukhada~USV.}
    \end{figure}

    
    The symmetrical properties of the Nukhada~USV entitles the execution of complex spiral manoeuvres. As detailed in~\cite{ittc2021tests}, the spiral manoeuvres is performed keeping the engine pods steering centred and actuating the thrusters in open-loop as shown in \fref{fig:nukhada_tests_spiral_propellers_cmd}. This is, initially, both thrusters are saturated at maximum forward power for $10$~seconds, thus making the catamaran to move forward. Then, for the next $50$~seconds, the setpoint of the left thruster is gradually changed from maximum forward power to maximum backward power; in this period, as the difference of thrust between propellers increases, the turning radius of the catamaran reduces all the way zero, i.e., the vehicle spinning over itself (see \fref{fig:nukhada_tests_spiral}).
    
    
    \section{CONCLUSION \label{sec:conclusion}}

In this manuscript, we have presented the Nukhada USV, an \acl{USV} for autonomous surveying and support to long-term underwater operations. The key features of the vehicle are incremental prototyping via its scaled version, highly customisable dry and wet payload with mission-specific equipment, and long-term autonomy. We have described the principal design aspects, as well as the composing subsystems and software architecture. Finally, we have detailed the payload the system is equipped with, and showed some of the trials conducted with the platform.

A total of seven \acp{USV} with slightly different platform design are currently under development for the \ac{MBZIRC}\footnote{\url{https://www.mbzirc.com/}}. Moreover, we are currently exploring the possibility of leveraging the spacing between the demi-hulls to include a docking and charging station to support different \acp{UUV} in long-term underwater robotic operations.
    
	\bibliographystyle{ieeetr}
    \bibliography{references}

\begin{thebibliography}{1}

\bibitem{ribas2011girona}
D.~Ribas, N.~Palomeras, P.~Ridao, M.~Carreras, and A.~Mallios, ``{Girona 500
  AUV: From survey to intervention},'' {\em IEEE/ASME Transactions on
  mechatronics}, vol.~17, no.~1, pp.~46--53, 2011.

\bibitem{carreras2015testing}
M.~Carreras, C.~Candela, D.~Ribas, N.~Palomeras, L.~Mag{\'\i}, A.~Mallios,
  E.~Vidal, {\`E}.~Vidal, and P.~Ridao, ``{Testing SPARUS II AUV, an open
  platform for industrial, scientific and academic applications},'' {\em
  Instrumentation viewpoint}, no.~18, pp.~54--55, 2015.

\bibitem{willners2021from}
J.~S. Willners, I.~Carlucho, T.~Łuczyński, S.~Katagiri, C.~Lemoine, J.~Roe,
  D.~Stephens, S.~Xu, Y.~Carreno, {\`E}.~Pairet, C.~Barbalata, Y.~Petillot, and
  S.~Wang, ``{From market-ready ROVs to low-cost AUVs},'' in {\em OCEANS
  2021-San Diego}, IEEE, 2021.

\bibitem{neira2021review}
J.~Neira, C.~Sequeiros, R.~Huamani, E.~Machaca, P.~Fonseca, and W.~Nina,
  ``Review on unmanned underwater robotics, structure designs, materials,
  sensors, actuators, and navigation control,'' {\em Journal of Robotics},
  vol.~2021, 2021.

\bibitem{quigley2009ros}
M.~Quigley, K.~Conley, B.~Gerkey, J.~Faust, T.~Foote, J.~Leibs, R.~Wheeler,
  A.~Y. Ng, {\em et~al.}, ``{ROS: an open-source Robot Operating System},'' in
  {\em ICRA workshop on open source software}, vol.~3, p.~5, Kobe, Japan, 2009.

\bibitem{coverageplanner}
R.~B{\"{a}}hnemann, N.~R.~J. Lawrance, J.~J. Chung, M.~Pantic, R.~Siegwart, and
  J.~I. Nieto, ``Revisiting boustrophedon coverage path planning as a
  generalized traveling salesman problem,'' {\em CoRR}, vol.~abs/1907.09224,
  2019.

\bibitem{ittc2021tests}
{International Towing Tank Conference}, ``Recommended procedures and guidelines
  - free running model tests.''
  \url{https://www.ittc.info/media/9681/75-02-06-01.pdf}.

\end{thebibliography}
\end{document}